\documentclass[preprint,11pt,authoryear]{elsarticle}

\usepackage{amsmath,amssymb}
\usepackage{booktabs}
\usepackage{adjustbox}
\usepackage{multirow}
\usepackage{longtable}
\usepackage{graphicx}
\usepackage{xcolor}
\usepackage[ruled,vlined,linesnumbered]{algorithm2e}
\usepackage{pgfplots}
\pgfplotsset{compat=1.16}
\usepackage[hidelinks]{hyperref}
\usepackage{orcidlink}

\journal{European Journal of Operational Research}

\begin{document}

\begin{frontmatter}

\title{A steady-state genetic algorithm with multi-ratio seeding and pair exchange for the quadratic knapsack problem}

\author[inst1]{Gaurav Srivastava\orcidlink{0000-0001-8675-138X}\corref{cor1}}
\ead{gauravs@iiita.ac.in}
\author[inst1]{Rahul Verma\orcidlink{0009-0002-8373-364X}}
\ead{pmm2025003@iiita.ac.in}
\author[inst2]{Alok Singh\orcidlink{0000-0003-3585-4957}}
\ead{alok@uohyd.ac.in}
\cortext[cor1]{Corresponding author}

\affiliation[inst1]{organization={Department of Information Technology, Indian Institute of Information Technology Allahabad},
            city={Prayagraj},
            postcode={211015},
            state={Uttar Pradesh},
            country={India}}
\affiliation[inst2]{organization={School of Computer and Information Sciences, University of Hyderabad},
            city={Hyderabad},
            postcode={500046},
            state={Telangana},
            country={India}}

\begin{abstract}
The quadratic knapsack problem (QKP) seeks a subset of items of maximum total linear and pairwise profit whose total weight does not exceed a given capacity. Most QKP heuristics rank items by a profit-to-weight ratio. We show on benchmark instances that this ranking can steer the search away from the best known solutions, and that on an almost full knapsack a local optimum can be separated from a better solution by an exchange where every single swap loses profit. We propose SSGA-P, a steady-state genetic algorithm with pair exchange that addresses both effects. Its initial population contains greedy solutions built with a grid of exponents between 0.5 and 3 applied to the item weights, and its pair exchange move combines an improving swap violating the capacity with a second swap or removal that restores feasibility. Incremental contribution vectors keep the method practical up to 10{,}000 items. SSGA-P finds optimal solutions in every run on the 100 classical instances and reaches all best known values in 99.8\% of the runs on the 80 instances with 1000 and 2000 items and in 99.75\% on the 40 instances with 5000 and 6000 items, significantly above the 87.7\% of the iterated hyperplane exploration algorithm, improving one best known value. On the 144 Large-QKP instances of Hochbaum et al. (2025), with the same 120-second limit, it improves 20 best known values and matches the other 124. An ablation study shows that the multi-ratio seeding and the pair exchange account for most of this performance.
\end{abstract}


\end{frontmatter}

\section{Introduction}
\label{sec:intro}

The quadratic knapsack problem (QKP) is one among the well-known extensions of the 0--1 knapsack problem. We are given a set $N=\{1,\dots,n\}$ of items and a knapsack of capacity $W$. Each item $i$ has a positive integer weight $w_i$ and a profit $p_{ii}\ge 0$, and every pair of distinct items $i,j$ has a profit $p_{ij}=p_{ji}\ge 0$ which is earned only if both items are packed together in the knapsack. The aim is to select a subset of items whose total weight does not exceed $W$ and whose total profit, individual and pairwise, is maximum. With a binary variable $x_i$ equal to 1 if item $i$ is selected and 0 otherwise, the problem can be written as
\begin{equation}
\max \Big\{ \sum_{i\in N} p_{ii}x_i + \sum_{i<j} p_{ij}x_ix_j \;:\; \sum_{i\in N} w_ix_i \le W,\; x\in\{0,1\}^n \Big\}.
\label{eq:qkp}
\end{equation}
If all pairwise profits are zero, (\ref{eq:qkp}) is the ordinary knapsack problem. If all weights are equal to one and the individual profits are zero, it contains the question of whether a graph has a clique of a given size. The QKP is therefore NP-hard in the strong sense \citep{Caprara1999,Pisinger2007}. Unlike the linear case, it cannot be solved by a  dynamic programming approach over capacities, since \citet{Fomeni2014} showed that Bellman's principle of optimality does not hold for it.

The problem was introduced under its present name by \citet{Gallo1980}, who motivated it with three examples: the location of satellite stations under a budget so that the traffic between the chosen stations is maximum, the choice of rainfall measuring sites with little redundancy, and the detection of cliques in graphs. Later work used the same model for the location of railway stations and freight terminals, for maximum dispersion and maximum diversity problems, for team formation, and as a subproblem in graph partitioning; the surveys of \citet{Pisinger2007} and \citet{Cacchiani2022} and the introduction of \citet{Hochbaum2025} list many of these applications. In most of them, the pairwise term represents a benefit that appears only when two elements are chosen together, and this is precisely what makes the problem harder than the linear knapsack problem: the contribution  of an item to the overall profit is not fixed but depends on what other items are in the knapsack.

Exact algorithms for the QKP are mostly of the branch-and-bound type, and their progress has followed the progress in upper bounds \citep{Gallo1980,Chaillou1989,Billionnet1999,Caprara1999,Billionnet2004}. The Quadknap algorithm of \citet{Caprara1999} solved instances with up to 400 items, and the aggressive reduction of \citet{Pisinger2007disp} pushed the limit to about 1500 items, although, as noted by \citet{Hochbaum2025}, its code is no longer available. The more recent cut-and-branch method of \citet{Fomeni2022} solves standard instances with up to 800 items within five hours. Instances with several thousand items, which are now common in the experimental literature, remain out of reach of exact methods, and for them the best results are due to heuristics.

Among these, a few methods stand out. The greedy randomized adaptive search procedure (GRASP) and tabu search algorithm of \citet{Yang2013} was the first to be tested on instances with 1000 and 2000 items. The iterated hyperplane exploration algorithm (IHEA) of \citet{ChenHao2017} restricts the search to hyperplanes on which the number of selected items is fixed and explores each of them by tabu search; it set the best known values for 80 instances with 1000 and 2000 items and for 40 instances with 5000 and 6000 items. Two recent papers approached the problem from quite different directions. \citet{Fennich2024} proposed a dynamic programming heuristic whose value function also estimates the future contribution of an item, and obtained near-optimal solutions on the hidden clique and dispersion instances that had been hard for earlier heuristics \citep{Schauer2016}. \citet{Hochbaum2025} proposed a breakpoints algorithm based approach for QKP named QKBP, which obtains optimal solutions for a set of budgets from the breakpoints of a parametric minimum cut and fills the remaining budgets greedily. They also compared eight methods under a common time limit of 120 seconds on more than a thousand instances, including a new Large-QKP collection with up to 10{,}000 items. In their study IHEA gave the best solutions on most instances, whereas QKBP was by far the fastest method and was close to the best values almost everywhere.

The present work grew out of a less ambitious attempt. We first applied a steady-state genetic algorithm with conventional operators, namely greedy initialization, crossover followed by greedy repair, and add, drop and swap local search, to the benchmark instances of \citet{Yang2013} and \citet{ChenHao2017}. The algorithm found the optimal solution for each and every small instance quickly, but it failed to reach the best known solutions on several of the 80 instances with 1000 and 2000 items, no matter how long it was run. We examined these failures one by one in a systematic manner and observed two recurring causes. The first is related to the way items are ranked. The algorithm, like most QKP heuristics, ordered candidate items by their current contribution divided by their weight. On some instances, the best known solution contains a mix of heavy and light items that this ordering never produces, whereas a greedy procedure that divides the contribution by $w_i^{\beta}$, for a value of $\beta$ different from one, reaches the best known value directly. The second cause arises when the knapsack is nearly full. A solution may then be locally optimal for every single add, drop or swap, and yet be improved by exchanging two selected items with two unselected ones, because one of the two swaps involved is improving but infeasible, and the other is feasible but not improving. Neither effect is specific to a genetic algorithm. Ranking by profit per unit of weight is used, in one form or another, in the initial construction and in the variable fixing of IHEA, in the greedy completion step of QKBP and in the local search of most other heuristics, and single-item neighbourhoods are the standard ones in the literature \citep{ChenHao2017}.

These observations motivated us to keep the simple steady-state framework, which has already proved effective for several combinatorial problems when its variation operators are designed considering the characteristics of the problem under consideration \citep{Srivastava2023,Srivastava2023eaai}, and to add components aimed at mitigating the two causes of failures that we have identified. The resulting algorithm, called SSGA-P (steady-state genetic algorithm with pair exchange), seeds its initial population with greedy solutions built for a grid of exponents $\beta\in[0.5,3]$, and its local search includes a pair exchange move that combines an improving swap that exceeds the capacity with a second swap or a removal that restores feasibility. SSGA-P differs from IHEA in that it does not fix the number of selected items and does not use tabu memory; diversity comes from the population and from occasional perturbation of the best solution. It differs from QKBP in that it is a stochastic search that spends its time budget improving solutions, rather than a one-pass construction. It also differs from the earlier genetic algorithm of \citet{Julstrom2005}, which used a single density ordering inside its operators and was tested on instances with up to 200 items.

The main contributions of the paper are the following.
\begin{enumerate}
\item We describe two causes of failure of ratio-based local search for the QKP, which we call ratio bias and budget-tight local optima, and illustrate them on benchmark instances.
\item We propose a multi-ratio seeding of the initial population and a pair exchange neighbourhood restricted to capacity classes, and embed them, together with an evict-and-insert move and a row-ordered update of the contribution vector, in a steady-state genetic algorithm that is practical up to 10{,}000 items.
\item We test the algorithm on 364 benchmark instances with 100 to 10{,}000 items. SSGA-P reaches all best known values on the three classical groups, is significantly more reliable than IHEA on the instances with 5000 and 6000 items, and improves the best known value of one of them. On the Large-QKP collection it finds better solutions than the eight methods compared by \citet{Hochbaum2025} on 20 instances and equal ones on the other 124.
\item An ablation study measures the contribution of each component and shows that the multi-ratio seeding and the pair exchange account for most of the improvement.
\end{enumerate}

The remainder of the paper is organised as follows. Section~\ref{sec:lit} reviews the literature on heuristics for the QKP in more detail. Section~\ref{sec:analysis} illustrates the two failure modes. Section~\ref{sec:alg} describes SSGA-P. Section~\ref{sec:exp} presents the computational study, and Section~\ref{sec:concl} provides the concluding remarks.

\section{Related literature}
\label{sec:lit}

The QKP has been reviewed by \citet{Pisinger2007} and, more recently, as part of the survey of knapsack problems by \citet{Cacchiani2022}. Both cover upper bounds and exact algorithms at length, and we only summarise them in Section~\ref{sec:lit-exact}. The remainder of this section focuses on heuristic methods, which serve as the relevant comparators for the present work.

\subsection{Upper bounds and exact algorithms}
\label{sec:lit-exact}

\citet{Gallo1980} proposed upper planes, linear functions that dominate the quadratic objective, and used both of them to bound the problem and to build an initial solution. \citet{Chaillou1989} derived a bound from a Lagrangian relaxation of the capacity constraint that can be computed with network flow techniques. \citet{Caprara1999} combined upper planes with Lagrangian relaxation in their Quadknap algorithm, and \citet{Billionnet1999} and \citet{Billionnet2004} obtained tighter bounds by Lagrangian decomposition. \citet{Pisinger2007disp} used these bounds to fix a large number of variables before branching, which made instances with up to about 1500 items solvable. \citet{Fomeni2022} developed a cut-and-branch method based on a strengthened linearisation. Despite this progress, the size of instances that can be solved to optimality has grown slowly, and none of these methods is applicable to the instances with 5000 or more items considered here.

\subsection{Constructive heuristics and metaheuristics}
\label{sec:lit-heur}

Early heuristics were constructive. The fill-up and exchange procedure of \citet{Gallo1980} fills the knapsack from an upper-plane solution and then improves it by swaps. \citet{Chaillou1989} proposed a greedy method that starts with all items and removes the item with the smallest ratio of contribution to weight until the capacity is respected. \citet{Julstrom2005} compared greedy, genetic and greedy-genetic algorithms. His best variant used the ratio of contribution to weight inside the crossover and mutation operators, and it outperformed a plain genetic algorithm on instances with up to 200 items. \citet{Hochbaum2025} re-implemented a multi-start version of this greedy method as one of their comparators. \citet{Singh2007} developed a hybrid steady-state genetic algorithm (HSSGA) in which infeasible offspring are repaired by a two-phase greedy heuristic based on relative value densities, and it outperformed the methods of \citet{Julstrom2005}. \citet{Xie2007} proposed a mini-swarm approach for the same small instances, and \citet{Pulikanti2009} combined an artificial bee colony algorithm with the greedy repair of HSSGA and a swap-based local search, obtaining results comparable to HSSGA and better than the mini-swarm. All these methods were evaluated on the classical instances with 100 and 200 items, and all of them rank items by profit per unit of weight.

\citet{Yang2013} combined a greedy randomised adaptive search procedure \citep{Feo1995} with tabu search. Solutions are built by a randomised greedy procedure and then improved by a tabu search over add, drop and swap moves. They introduced the 80 instances with 1000 and 2000 items that we call Group~II, and reported a success rate of 91.6\% over all runs.

The iterated hyperplane exploration algorithm of \citet{ChenHao2017} is, to our knowledge, the strongest published metaheuristic for the standard instances. It starts from the observation that the number of items in an optimal solution lies between a lower bound $k_{LB}$ and an upper bound $k_{UB}$ which are easy to compute, and that near-optimal solutions tend to have close to $k_{UB}$ items. For each candidate value of $k$, it adds the hyperplane $\sum_i x_i=k$ to the model, fixes to one a number of selected items with the largest density, defined as contribution divided by weight, and solves the reduced problem with a tabu search that may pass through infeasible solutions. A perturbation step moves the search to other hyperplanes when no further improvement is found. With 100 runs per instance and a limit of $n+65$ hyperplane iterations, IHEA reached the best known values in 99.9\% of the runs on Group~II, and in 87.7\% of the runs on the 40 instances with 5000 and 6000 items (Group~III) that it introduced. IHEA is an elaborate method with several interacting components and parameters, which makes it hard to determine which of them is responsible for its performance.

\subsection{Dynamic programming heuristics}
\label{sec:lit-dp}

\citet{Fomeni2014} showed that the dynamic programme for the linear knapsack problem does not give an exact algorithm for the QKP, but that it gives a good heuristic if the value of each state is computed with respect to the items already packed. \citet{Fomeni2023} later applied the same idea in the space of lifted quadratic variables. \citet{Fennich2024} improved the value function further, so that it also takes into account the expected contribution of an item with respect to items that have not yet been considered, and added a propagation step and a remove-and-fill-up local search. Following \citet{Schauer2016}, who showed that the classical random instances become easy for simple greedy methods as $n$ grows, they evaluated their method mainly on structured instances derived from dispersion and densest subgraph problems and on hidden clique instances, where it was much better than earlier heuristics. Their study is a reminder that good performance on the classical random instances alone may hide weaknesses. It is also deterministic and relatively fast, but it was not tested on instances with thousands of items.

\subsection{The breakpoints approach and the 2025 comparison}
\label{sec:lit-qkbp}

\citet{Hochbaum2025} represented the QKP on a directed graph and relaxed the capacity constraint with a Lagrange multiplier $\lambda$. For every $\lambda$, the relaxed problem is a minimum cut problem, and the solutions for all values of $\lambda$ are nested and can be obtained together by a parametric minimum cut computation. Each breakpoint of the resulting concave envelope is an optimal solution for its own budget. For a budget that falls between two breakpoints, QKBP adds items greedily to the smaller breakpoint solution or removes items greedily from the larger one, and keeps the better result. The method is deterministic and very fast: it solves instances with 10{,}000 items in seconds.

The computational study in that paper is the most extensive comparison of QKP methods to date. Eight methods were compared: QKBP, the relative greedy method based on \citet{Julstrom2005}, IHEA, the dynamic programming heuristics of \citet{Fomeni2014} and \citet{Fomeni2023}, the exact method Quadknap, and the commercial solvers Gurobi and Hexaly, each run once per instance with a limit of 120 seconds. Besides the classical collections, the study introduced the Large-QKP collection of 144 instances with 500 to 10{,}000 items, densities from 5\% to 100\% and six budget levels. IHEA obtained the best value on most instances, but its running time grew quickly with $n$; QKBP was within a fraction of a percent of the best value on almost all instances in a small part of the time. We are not aware of later work that improves on the values reported in that study.

\subsection{Steady-state genetic and memetic algorithms}
\label{sec:lit-ssga}

In a steady-state genetic algorithm only one or a few worst individuals are replaced in each generation \citep{Whitley1989}, so that good solutions are kept in the population and are immediately available for further recombination. When each offspring is improved by local search, the method belongs to the family of memetic algorithms \citep{Neri2012}. The steady-state model has been applied with success to grouping, scheduling and routing problems in which the crossover and mutation operators are designed according to the structure of the problem, and the initial population is built with a mix of greedy and random heuristics \citep{Srivastava2023,Srivastava2023eaai}. For the QKP itself, a steady-state genetic algorithm (HSSGA) was first used by \citet{Singh2007}. Their HSSGA uses the standard binary encoding, builds the child as the union of the two parents, applies bit-flip mutation, and repairs the resulting infeasible chromosome with a greedy heuristic based on relative value densities; it was tested on 20 instances with 100 and 200 items. SSGA-P uses the same binary encoding and the steady-state population replacement model, but  other key components are altogether different: crossover keeps only the items common to both the parents and fills the remaining capacity semi-greedily, so that every offspring is feasible and no repair is needed; mutation removes and replaces a few items; offspring may be subjected  to a local search strategy comprising several types of moves; and the multi-ratio seeding and the pair exchange are new. SSGA-P is also designed for instances with up to 10{,}000 items. Its perturbation of the best solution is borrowed from iterated local search \citep{Lourenco2003}, and its semi-greedy constructions use a restricted candidate list (RCL) in the spirit of GRASP \citep{Feo1995}.

To summarise, the literature offers a very strong but complex tabu-search metaheuristic (IHEA), a family of deterministic dynamic programming heuristics that are effective on structured instances, and a very fast deterministic method based on parametric cuts (QKBP). All of them rank items, at some stage, by contribution per unit of weight, and their improvement phases rely on single-item moves. In the following section, we show on concrete instances why this can be a limitation, which leads to the design of SSGA-P in Section~\ref{sec:alg}.

\section{Two failure modes of ratio-based search}
\label{sec:analysis}

Throughout the paper we write $S=\{i : x_i=1\}$ for the set of items in the current solution. The value of an item in the QKP is not fixed: besides its own profit $p_{ii}$, it earns a pairwise profit $p_{ij}$ with every other selected item $j$. We therefore define the \emph{contribution} of item $i$ with respect to $S$ as
\begin{equation}
c_i(S) = p_{ii} + \sum_{j\in S,\, j\ne i} p_{ij},
\label{eq:contrib}
\end{equation}
that is, the profit that item $i$ earns together with the items currently in $S$. The definition applies both to selected and to unselected items; for an unselected item it is the profit the item would bring if it were added. With this notation the effect of the three basic moves of a local search follows directly.

\begin{itemize}
\item \emph{Add} an item $j\notin S$. The new item brings its own profit and its pairwise profit with every item of $S$, so the objective changes by $c_j(S)$.
\item \emph{Drop} an item $i\in S$. The solution loses the own profit of $i$ and all pairwise profits between $i$ and the remaining items, so the objective changes by $-c_i(S)$.
\item \emph{Swap} an item $i\in S$ for an item $j\notin S$. This is a drop followed by an add, but the two terms are not independent: $c_j(S)$ is computed while $i$ is still in $S$ and therefore contains $p_{ij}$, which is not earned once $i$ has left. The change is
\begin{equation}
\Delta_{\mathrm{swap}}(i,j) = c_j(S)-c_i(S)-p_{ij}.
\label{eq:swap}
\end{equation}
\end{itemize}

For example, with $p_{11}=10$, $p_{22}=6$, $p_{33}=8$, $p_{12}=4$, $p_{13}=2$, $p_{23}=5$ and $S=\{1,2\}$, the objective is $20$ and $c_1=14$, $c_2=10$, $c_3=15$. Adding item 3 gives $20+15=35$, dropping item 1 gives $20-14=6$, and swapping item 1 for item 3 gives $20+(15-14-2)=19$, which is the value of $\{2,3\}$; without the term $-p_{13}$ the swap would be overestimated by 2.

If the contributions were recomputed from (\ref{eq:contrib}) for every candidate move, each evaluation would cost $O(n)$. Instead, we store the vector $c=(c_1,\dots,c_n)$ for the current solution and evaluate every add, drop and swap with one or two look-ups, that is, in $O(1)$ time. The vector must be updated only when a move is actually applied: after item $k$ is added, $c_i \leftarrow c_i+p_{ik}$ for all $i\in N$, and after $k$ is dropped, $c_i \leftarrow c_i-p_{ik}$. Each update reads one row of the profit matrix and costs $O(n)$. Since a local search evaluates many more moves than it applies, this trade-off reduces the running time considerably, and it is what makes the algorithm more suitable for instances with $10{,}000$ items.

\subsection{Ratio bias}
\label{sec:ratio}

\begin{figure}[t]
\centering
\begin{tikzpicture}
\begin{axis}[width=0.48\textwidth,height=5.2cm,xlabel={$\beta$},ylabel={gap to best known (\%)},title={1000\_75\_5},xmin=0.5,xmax=3,ymin=0,grid=both,grid style={gray!20},tick label style={font=\footnotesize},label style={font=\small},title style={font=\small}]
\addplot[mark=*,mark size=1pt,blue] table[x=beta,y=gap]{figures/beta_1000_75_5.dat};
\end{axis}
\end{tikzpicture}\hfill
\begin{tikzpicture}
\begin{axis}[width=0.48\textwidth,height=5.2cm,xlabel={$\beta$},title={2000\_50\_5},xmin=0.5,xmax=3,ymin=0,grid=both,grid style={gray!20},tick label style={font=\footnotesize},label style={font=\small},title style={font=\small}]
\addplot[mark=*,mark size=1pt,red] table[x=beta,y=gap]{figures/beta_2000_50_5.dat};
\end{axis}
\end{tikzpicture}
\caption{Relative gap between a pure greedy construction ranked by $c_j/w_j^{\beta}$ and the best known value, for $\beta=0.50,0.55,\dots,3.00$. The usual ratio corresponds to $\beta=1$.}
\label{fig:beta}
\end{figure}
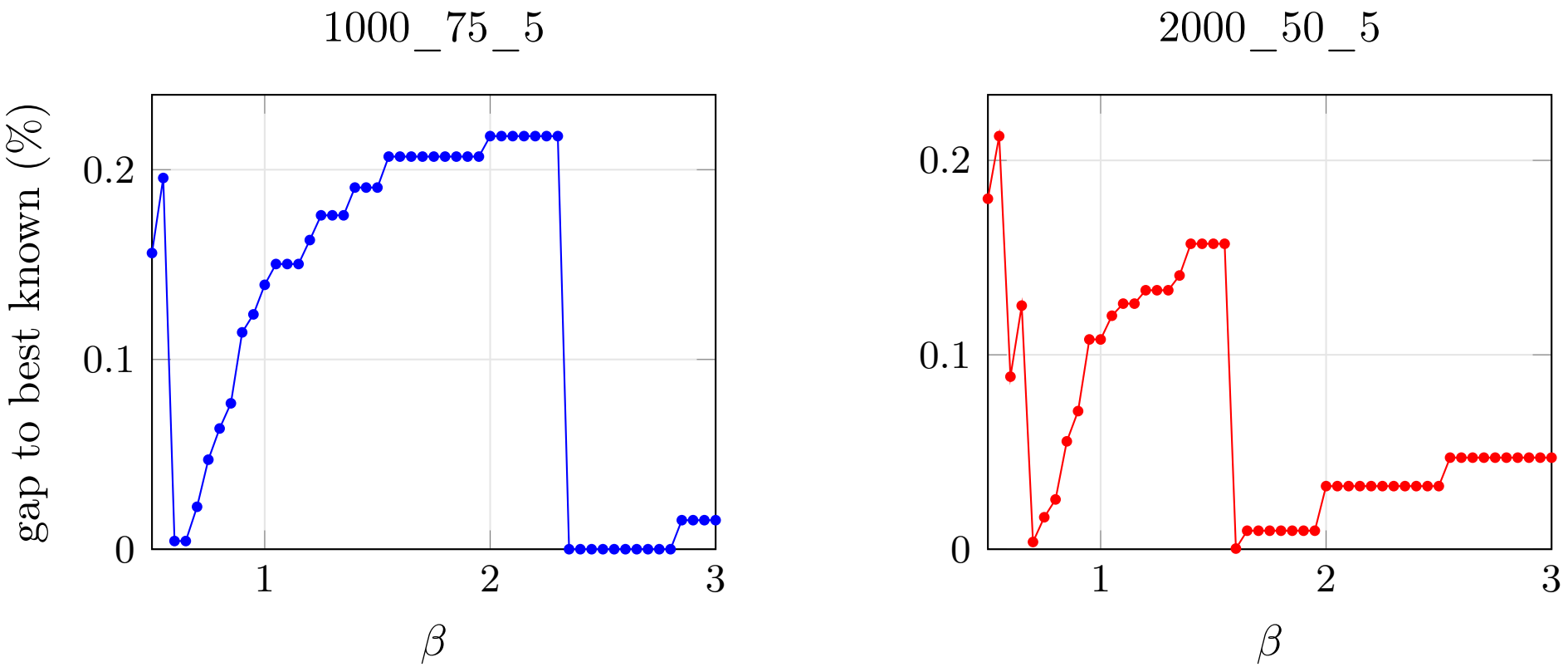

Greedy constructions, the refill steps of crossover and destroy-and-repair moves, and the item ordering in IHEA all rank the items $j\notin S$ that still fit by the ratio $c_j(S)/w_j$. Consider instead the family of rankings
\begin{equation}
r_j^{\beta}(S) = c_j(S) / w_j^{\beta}, \qquad \beta>0 .
\label{eq:beta}
\end{equation}
Values $\beta>1$ penalize weight more than the usual ratio and favour many light items; values $\beta<1$ favour heavier items with large contributions. Figure~\ref{fig:beta} shows the relative gap to the best known value of a pure greedy construction (no local search) with ranking~\eqref{eq:beta} on two instances of Group~II. On instance 1000\_75\_5, the usual ratio leaves a gap of 0.139\%, while every $\beta$ in $[2.35,2.80]$ gives the best known value 13{,}970{,}842 directly. On 2000\_50\_5 the gap at $\beta=1$ is 0.108\%, and a narrow window around $\beta=1.6$ brings the greedy solution within 0.0003\% of the best known value. The curves are far from monotone, so there is no single exponent that one could recommend. A steady-state genetic algorithm with the same operators as SSGA-P but without multi-ratio seeding never reached the best known value on either instance in 30 runs; it stopped at 13{,}970{,}240 and 16{,}805{,}999, respectively. Interestingly, 13{,}970{,}240 is exactly the value of the greedy solution for $\beta=0.6$: the search settled in a basin that one particular ranking leads to.

\subsection{Budget-tight local optima}
\label{sec:tight}

The second effect appears when a solution uses the whole capacity. Table~\ref{tab:pair} shows a local optimum of instance 5000\_50\_4 of Group~III, with value 170{,}999{,}968, that is 260 below the best known value (BKV), and zero residual capacity, that differs from the best known solution in three items on each side. None of the one-for-one swaps between these items improves the solution: two lose profit and the third does not fit, because the entering item is one unit heavier than the leaving one. The exchange of two items for two items is feasible and gains 223, and the full three-for-three exchange gains 260. In our runs, add/swap descent, random destroy-and-repair and evict-and-insert (Section~\ref{sec:ls}) all failed to leave this solution, which was the final solution in three of seven runs of an earlier version of the algorithm without pair exchange.

\begin{table}[t]
\centering
\caption{Exchanges between a local optimum and the best known solution of instance 5000\_50\_4 (item indices start at 1). The local optimum uses the capacity $W=69{,}485$ exactly.}
\label{tab:pair}
\begin{adjustbox}{max width=\linewidth}
\begin{tabular}{lcr}
\toprule
Exchange (leave $\rightarrow$ enter) & Weight & Change in profit\\
\midrule
$\{3380\}\rightarrow\{413\}$ & 39 $\rightarrow$ 38 & $-1{,}782$ \\
$\{3324\}\rightarrow\{3125\}$ & 38 $\rightarrow$ 38 & $-19$ \\
$\{35\}\rightarrow\{3621\}$ & 36 $\rightarrow$ 37 & infeasible (+1) \\
$\{35,3380\}\rightarrow\{413,3621\}$ & 75 $\rightarrow$ 75 & $+223$ \\
$\{35,3324,3380\}\rightarrow\{413,3125,3621\}$ & 113 $\rightarrow$ 113 & $+260$ \\

\bottomrule
\end{tabular}
\end{adjustbox}
\end{table}

This configuration is not specific to one instance. When good solutions use almost all the capacity, the weights of entering and leaving items must match closely, and an improving move may have to borrow a few units of capacity from a second, slightly unprofitable exchange. The pair exchange of Section~\ref{sec:pair} searches exactly for such combinations.

\section{The SSGA-P algorithm}
\label{sec:alg}

SSGA-P is a steady-state genetic algorithm with a population of $\mu=100$ distinct feasible solutions kept sorted by objective value. Algorithm~\ref{alg:ssgap} gives the overall scheme. The parameter values are listed in Table~\ref{tab:param}; they were fixed during development on a small number of Group~II instances and were not tuned per collection.

\begin{algorithm}[t]
\caption{SSGA-P}
\label{alg:ssgap}
\KwIn{instance $(P,w,W)$, time limit $T$}
\KwOut{best solution found}
$\mathcal{P}\leftarrow$ \textsc{InitPopulation}() \tcp*{Section~\ref{sec:init}}
\While{elapsed time $<T$}{
  $A,B\leftarrow$ two binary tournaments on $\mathcal{P}$\;
  $C\leftarrow$ \textsc{Crossover}$(A,B)$; $M\leftarrow$ \textsc{Mutate}$(C)$; $X\leftarrow$ the better of $C$ and $M$\;
  with probability 0.2 apply add/swap descent to $X$, otherwise with probability 0.3 apply random drop-add\;
  with probability 0.2 apply evict-and-insert to $X$\;
  replace the worst member of $\mathcal{P}$ by $X$ if $X$ is better and not already in $\mathcal{P}$\;
  \If{no new best solution for $g_{\max}$ generations}{
     every third time, rebuild the worse half of $\mathcal{P}$\;
     insert \textsc{Kick}(best solution) into $\mathcal{P}$ \tcp*{Section~\ref{sec:kick}}
  }
}
apply full descent with pair exchange to the best solution and return it\;
\end{algorithm}

\begin{table}[t]
\centering
\caption{Parameters of SSGA-P (identical for all collections).}
\label{tab:param}
\small
\begin{adjustbox}{max width=\linewidth}
\begin{tabular}{lp{0.72\textwidth}}
\toprule
Component & Setting\\
\midrule
Population & $\mu=100$, binary tournament, replace worst, no duplicates\\
Seeding & one greedy solution with $\beta=1$; greedy for $\beta=0.50,0.55,\dots,3.00$ (51 solutions), each improved by add/swap local search; 20 semi-greedy solutions (RCL (restricted candidate list) parameter 0.3, random $\beta\in[0.5,3]$); 28 random solutions\\
Crossover & common core, semi-greedy fill with RCL parameter 0.2\\
Mutation & drop 1--3 random items, refill with random fitting items; the better of child and mutant is kept\\
Local search & per offspring: add/swap (prob.\ 0.2) or random drop-add of 2--5\% of $S$ (prob.\ 0.3);\\
 & evict-and-insert (prob.\ 0.2, 10 tries)\\
Pair exchange & 64 best candidates per capacity class, classes up to 64 units, excess $e$ matched with freed capacity $f\in[e,e+3]$;\\
 & applied after each kick and in the final descent\\
Perturbation & after $g_{\max}=\max(200,n/2)$ generations without improvement: drop 10--20\% of the best solution, refill semi-greedily (RCL parameter 0.1) with random $\beta$, then local search;\\
 & every third kick also rebuilds the worse half of the population\\
Final descent & at the time limit, add/swap, evict-and-insert and pair exchange are repeated on the best solution until none improves it\\
Time limit & Groups I--II: $\max(1, 6n/1000)$ s; Group III: $18n/1000$ s; Large-QKP: 120 s\\
\bottomrule
\end{tabular}
\end{adjustbox}
\end{table}

\subsection{Solution representation and fitness}
\label{sec:repr}

Each individual of the population is represented by a binary chromosome $x=(x_1,\dots,x_n)$ of length $n$, in which gene $x_i=1$ if item $i$ is packed and $x_i=0$ otherwise; the set of selected items is therefore $S=\{i : x_i=1\}$. This direct encoding is the one commonly used for knapsack problems \citep{Julstrom2005,Singh2007} and needs no decoding step. All operators of SSGA-P generate only feasible chromosomes: items are added only when they fit into the remaining capacity, so the total weight $\sum_{i\in S} w_i$ never exceeds $W$ and no penalty function or repair of infeasible individuals is required. The fitness of an individual is its objective value~\eqref{eq:qkp}. Together with the chromosome, each individual stores its profit and total weight, which are updated incrementally when items are added or removed, and the population contains no two identical chromosomes.

\subsection{Greedy fill and multi-ratio seeding}
\label{sec:init}

All constructive steps use the same procedure. Given a partial solution $S$, an optional set of allowed items and a parameter $\alpha\in[0,1]$, the fill repeatedly computes $r_j^{\beta}(S)$ for every allowed item $j\notin S$ that fits and has $c_j(S)>0$, forms the restricted candidate list $\{j : r_j^{\beta}\ge r_{\max}-\alpha(r_{\max}-r_{\min})\}$, and adds a random member of that list. With $\alpha=0$ the fill is the pure greedy construction of Section~\ref{sec:ratio}.

The initial population of 100 solutions is built in four parts. The first solution is the greedy solution with the usual ratio, $\beta=1$, without application of local search. The next 51 are greedy solutions for $\beta=0.50,0.55,\dots,3.00$, each improved by the add/swap local search of Section~\ref{sec:ls}. They are followed by 20 semi-greedy solutions, in which each item is chosen at random from a restricted candidate list ($\alpha=0.3$) and the exponent is drawn uniformly from $[0.5,3]$. The remaining 28 solutions are random: items are taken in random order and added whenever they fit.

Every new solution is compared with those already in the population, and duplicates are not accepted. A semi-greedy or random solution that duplicates an existing one is simply generated again. A grid solution cannot be regenerated, since the same exponent always gives the same solution; when it duplicates one already present, which happens often because neighbouring exponents produce the same ordering, the slot receives a random solution instead.

The grid is the only place where all exponents are tried systematically. Afterwards, random exponents are used in restarts and kicks, while crossover, mutation and the local search rank items with $\beta=1$.  

\subsection{Crossover and mutation}

The crossover keeps the items common to both parents, which is feasible because it is a subset of either parent, and then fills the remaining capacity semi-greedily, first from the items that belong to exactly one parent and then from all items. The mutation removes one to three random items and refills the capacity with random items of positive contribution, excluding the items just removed. The better of the two offspring is passed to the local search.

\subsection{Local search moves}
\label{sec:ls}

\paragraph{Add/swap descent} The descent visits the items in random order and applies, with first improvement, all improving additions, and one-for-one swaps, followed by additions that the swaps have made possible. At most two passes are made inside the generation loop; the final descent repeats until no move improves.

\paragraph{Random drop-add} A random 2--5\% of the selected items are removed and the capacity is refilled greedily, first without the removed items and then with all items. The result is kept only if it is not worse. In preliminary tests, a random choice of the removed items worked as well as choices guided by efficiency.

\paragraph{Evict-and-insert} An unselected item $u$ with positive contribution is chosen by a tournament of size three on $c_u/w_u$. If it does not fit, the selected items are removed in increasing order of $(c_i - p_{iu})/w_i$, the profit lost per unit of capacity freed taking the interaction with $u$ into account, until $u$ fits; then $u$ is inserted and the capacity refilled. The best of ten attempts is accepted if it improves. The move allows a heavy item to replace several light ones.

\subsection{Pair exchange}
\label{sec:pair}

Let $\sigma=W-\sum_{i\in S}w_i$ be the residual capacity. For a swap $(i,j)$ with $i\in S$, $j\notin S$, let $\delta_{ij}=c_j-c_i-p_{ij}$ be its gain and $e_{ij}=w_j-w_i-\sigma$ the capacity it lacks. The pair exchange combines an \emph{improving infeasible} swap $(i_1,j_1)$ with $\delta_{i_1j_1}>0$ and $e=e_{i_1j_1}>0$ with a \emph{releasing} move that frees at least $e$ units: either a swap $(i_2,j_2)$ with $f=w_{i_2}-w_{j_2}\ge e$, or the removal of $i_2$ with $f=w_{i_2}$. The gain of the combined move is
\begin{equation}
\Delta = \delta_{i_1j_1} + \delta_{i_2j_2} + p_{i_1i_2} - p_{i_2j_1} + p_{j_1j_2} - p_{i_1j_2},
\label{eq:pairgain}
\end{equation}
where for a removal $\delta_{i_2j_2}$ is replaced by $-c_{i_2}$ and the last two terms vanish. One scan over all pairs $(i,j)$ fills, for every lacking amount $e\le 64$, a list of the 64 most profitable improving infeasible swaps, and for every released amount $f\le 64$ a list of the 64 most profitable releasing moves. The combination step evaluates~\eqref{eq:pairgain} for every pair from lists with $e\le f\le e+3$, applies the best combination if $\Delta>0$, and refills the capacity greedily. Grouping by capacity class is essential: among more than two million improving infeasible swaps of the local optimum of Table~\ref{tab:pair}, the useful one ranks only third among those lacking exactly two units. The scan costs $O(|S|\,(n-|S|))$ and the combination at most $64^3\cdot 4$ evaluations, so the move is used in the final descent and after each kick rather than in every generation.

The three numerical limits of the pair exchange are design choices; they were not tuned. Each of them can be justified as follows.

\emph{At most 64 units per capacity class.} A capacity class collects the moves that lack, or free, the same number of units. In all collections used here the item weights are integers between 1 and 50. A swap therefore changes the total weight by at most 49 units and a removal frees at most 50 units, so every possible class lies below 64. This limit therefore never excludes a move; it only sets the size of the lists in the implementation.

\emph{At most 64 moves per class.} Each class can contain thousands of moves, and combining all of them would be too slow. We therefore keep only the 64 moves with the largest gain in each class. Little is lost by this, because the gain of a combined move~\eqref{eq:pairgain} consists mainly of the two individual gains; the four remaining terms are single pairwise profits, which are small in comparison. A combination of two moves that are both poor in their classes is therefore unlikely to be the best one. In the example of Table~\ref{tab:pair}, the useful swap ranks third in its class.

\emph{The released capacity lies between $e$ and $e+3$.} The releasing move must free at least the $e$ units that the first swap lacks, otherwise the combined move is still infeasible. It should not free much more, because on a full knapsack every unused unit of capacity is lost profit. We therefore allow up to three extra units, which are then used, if possible, by the greedy filling that follows the move.

\subsection{Perturbation}
\label{sec:kick}

If no new best solution is found for $g_{\max}$ generations, the best solution is kicked: 10--20\% of its items are removed at random, the capacity is refilled semi-greedily with a random exponent, excluding the removed items first, and the result is improved by descent, evict-and-insert and pair exchange. The kicked solution replaces the worst member of the population even if it is worse. Every third kick the worse half of the population is rebuilt with semi-greedy and random solutions.

\subsection{Implementation}

The contribution vector is updated after each addition or removal of item $k$ by reading row $k$ of $P$, which is the same as column $k$ because $P$ is symmetric but is contiguous in memory. For $n>2000$ the full vector is recomputed by adding the rows of the selected items. With $P$ stored as a dense matrix of 4-byte integers (400 MB for $n=10{,}000$), these two changes alone increased the number of generations per second by a factor of about four at $n=6000$. The program is written in {\sc C}, single-threaded, and uses no external solver.

\section{Computational study}
\label{sec:exp}

\subsection{Instances, setting and reference results}

We use four collections: Group~I, the 100 classical instances with 100 to 300 items \citep{Billionnet2004}, for which optimal values are known; Group~II, 80 instances with 1000 and 2000 items \citep{Yang2013}; Group~III, 40 instances with 5000 and 6000 items \citep{ChenHao2017}; and Large-QKP, 24 graphs with 500 to 10{,}000 items and densities from 5\% to 100\%, each with six capacities equal to 2.5, 5, 10, 25, 50 and 75\% of the total weight, giving 144 instances \citep{Hochbaum2025}. Large-QKP instances were generated with the authors' published procedure and seed; the edge counts agree with those reported by the authors.

All experiments were run on a computer with an Intel Core i7-3770 processor (3.4~GHz) and 12~GB of memory under Ubuntu, with one run per core, and the code was compiled with gcc using \texttt{-O2}. Every run uses a fixed CPU time limit (Table~\ref{tab:param}); time-to-best is the CPU time at which the final best solution was first found. Each Group~I--III instance was solved 30 times and each Large-QKP instance 10 times. A run is a \emph{hit} if it reaches the best known value (BKV).

For Groups~II and~III the BKVs and the results of IHEA and GRASP+tabu are taken from \citet{ChenHao2017}, where each method was run 100 times; IHEA was run on an AMD Opteron 4184 at 2.8~GHz. We also report the results of QKBP and of IHEA from the comparison of \citet{Hochbaum2025}, in which every method was run once with a limit of 120~s on an Intel Xeon at 3.3~GHz. Running times on different machines are given for orientation only.

\subsection{Group I}

Table~\ref{tab:g1} summarizes the results. SSGA-P finds the optimal value of all 100 instances in all 3000 runs, in 0.02~s on average. The comparison is not informative here: IHEA and GRASP+tabu also solve nearly all runs to optimality \citep{ChenHao2017}, and the instances mainly serve as a check.

\begin{table}[t]
\centering
\caption{Group I: number of instances, instances whose optimum was found, percentage of runs reaching the optimum (hits), average time to best and average run time (s).}
\label{tab:g1}
\small
\begin{adjustbox}{max width=\linewidth}
\begin{tabular}{rrrrrrr}
\toprule
$n$ & Density (\%) & Inst. & Optimum found & Hits (\%) & Time to best & Run time\\
\midrule
100 & 25 & 10 & 10 & 100.0 & 0.009 & 1.00 \\
100 & 50 & 10 & 10 & 100.0 & 0.005 & 1.00 \\
100 & 75 & 10 & 10 & 100.0 & 0.011 & 1.00 \\
100 & 100 & 10 & 10 & 100.0 & 0.009 & 1.00 \\
200 & 25 & 10 & 10 & 100.0 & 0.030 & 1.21 \\
200 & 50 & 10 & 10 & 100.0 & 0.030 & 1.20 \\
200 & 75 & 10 & 10 & 100.0 & 0.014 & 1.20 \\
200 & 100 & 10 & 10 & 100.0 & 0.017 & 1.20 \\
300 & 25 & 10 & 10 & 100.0 & 0.036 & 1.80 \\
300 & 50 & 10 & 10 & 100.0 & 0.034 & 1.80 \\

\bottomrule
\end{tabular}
\end{adjustbox}
\end{table}

\subsection{Group II}

Table~\ref{tab:g2} reports averages per class of 10 instances; per-instance results are in~\ref{app:inst}. SSGA-P reaches the BKV of all 80 instances, in 2395 of the 2400 runs (99.8\%), with every run successful on 78 instances. The only exceptions are 1000\_50\_5 (29 of 30 runs) and 1000\_50\_8 (26 of 30). IHEA reports a success rate of 99.9\% over 100 runs and GRASP+tabu 91.6\%. A Wilcoxon signed-rank test \citep{Wilcoxon1945,Derrac2011} on the per-instance success rates finds no significant difference between SSGA-P and IHEA ($p=0.59$). In the single runs of \citet{Hochbaum2025}, QKBP is on average 0.04--0.23\% below the BKV, depending on the class, and IHEA up to 0.011\%. SSGA-P finds the best solution after 1.4~s on average, within a limit of 6 or 12~s.

\begin{table}[t]
\centering
\caption{Group II, averages over the 10 instances of each class. SSGA-P: instances whose BKV was reached, hits (\%), average gap to the BKV (\%), time to best and run time (s). IHEA and GRASP+tabu: success rate (\%) and IHEA time (s) from \citet{ChenHao2017}. Last two columns: average gap (\%) of the single runs of IHEA and QKBP in \citet{Hochbaum2025}.}
\label{tab:g2}
\scriptsize
\setlength{\tabcolsep}{4pt}
\begin{adjustbox}{max width=\linewidth}
\begin{tabular}{rr rrrrr rrr rr}
\toprule
 & & \multicolumn{5}{c}{SSGA-P (30 runs)} & \multicolumn{3}{c}{\citet{ChenHao2017}} & \multicolumn{2}{c}{\citet{Hochbaum2025}}\\
\cmidrule(lr){3-7}\cmidrule(lr){8-10}\cmidrule(lr){11-12}
$n$ & Dens. & BKV & Hits & Gap & $t_{\mathrm{best}}$ & $t$ & IHEA SR & IHEA $t$ & GRASP SR & IHEA gap & QKBP gap\\
\midrule
1000 & 25 & 10/10 & 100.0 & 0.00000 & 0.47 & 6.0 & 100.0 & 5.5 & 96.6 & 0.011 & 0.209 \\
1000 & 50 & 10/10 & 98.3 & 0.00000 & 1.04 & 6.0 & 99.7 & 6.0 & 88.4 & 0.002 & 0.074 \\
1000 & 75 & 10/10 & 100.0 & 0.00000 & 0.59 & 6.0 & 100.0 & 6.4 & 95.8 & 0.000 & 0.232 \\
1000 & 100 & 10/10 & 100.0 & 0.00000 & 0.80 & 6.0 & 100.0 & 6.1 & 95.6 & 0.000 & 0.184 \\
2000 & 25 & 10/10 & 100.0 & 0.00000 & 1.99 & 12.1 & 100.0 & 22.2 & 95.6 & 0.001 & 0.040 \\
2000 & 50 & 10/10 & 100.0 & 0.00000 & 2.16 & 12.1 & 100.0 & 23.2 & 78.5 & 0.001 & 0.072 \\
2000 & 75 & 10/10 & 100.0 & 0.00000 & 1.52 & 12.1 & 100.0 & 22.7 & 93.7 & 0.000 & 0.066 \\
2000 & 100 & 10/10 & 100.0 & 0.00000 & 2.71 & 12.1 & 99.5 & 22.7 & 88.8 & 0.005 & 0.096 \\

\bottomrule
\end{tabular}
\end{adjustbox}
\end{table}

\subsection{Group III}

Table~\ref{tab:g3} gives the results for the 40 largest instances of \citet{ChenHao2017}. With a limit of $18n/1000$ seconds, that is 90~s for $n=5000$ and 108~s for $n=6000$, SSGA-P reaches the BKV of all 40 instances in 1197 of 1200 runs (99.75\%). The three unsuccessful runs concern 5000\_25\_5, where they end 0.0001\% below the BKV. IHEA reaches the BKVs in 87.7\% of its runs, with success rates of 1\% on 6000\_100\_2, 4\% on 5000\_75\_5 and 12\% on 6000\_75\_1. SSGA-P has a higher success rate than IHEA on 10 instances and a lower one on one; the difference is significant ($p=0.006$). For instance 6000\_75\_1, every run of SSGA-P returns a solution of value 204{,}528{,}201, which improves the BKV 204{,}512{,}250 by 15{,}951; the solution is given in the supplementary material. To our knowledge no better value has been published; the single IHEA run of \citet{Hochbaum2025} also returned 204{,}512{,}250.

The 2025 comparison also shows that a single IHEA run does not reach the BKV of 5000\_100\_5, 5000\_75\_5 and 6000\_100\_2 within its time limit. Its runs took on average 154~s for $n=5000$ and 169~s for $n=6000$, longer than the nominal 120~s, because the time is checked only between iterations. On two instances, 6000\_100\_2 and 6000\_100\_5, SSGA-P obtains the BKV in the final descent with pair exchange at the end of each run; we return to this point in Section~\ref{sec:ablation}.

The time limit for Group~III is three times the one used for Group~II. With $6n/1000$ seconds, the success rate was 73.6\% (with an earlier implementation without the row-ordered updates and without pair exchange), and on several instances the time to best was close to the limit. Even with the larger limit, SSGA-P uses less time per run than IHEA in both studies, after allowing for the different processors.

\begin{table}[t]
\centering
\caption{Group III, averages over the 5 instances of each class; columns as in Table~\ref{tab:g2}. GRASP+tabu was not run on these instances. The negative average gap for $n=6000$, density 75\% reflects the improved BKV of 6000\_75\_1.}
\label{tab:g3}
\scriptsize
\setlength{\tabcolsep}{4pt}
\begin{adjustbox}{max width=\linewidth}
\begin{tabular}{rr rrrrr rr rr}
\toprule
 & & \multicolumn{5}{c}{SSGA-P (30 runs)} & \multicolumn{2}{c}{IHEA (100 runs)} & \multicolumn{2}{c}{\citet{Hochbaum2025}}\\
\cmidrule(lr){3-7}\cmidrule(lr){8-9}\cmidrule(lr){10-11}
$n$ & Dens. & BKV & Hits & Gap & $t_{\mathrm{best}}$ & $t$ & SR & $t$ & IHEA gap & QKBP gap\\
\midrule
5000 & 25 & 5/5 & 98.0 & 0.00002 & 15.08 & 90.4 & 100.0 & 130.4 & 0.000 & 0.028 \\
5000 & 50 & 5/5 & 100.0 & 0.00000 & 37.88 & 90.4 & 100.0 & 128.0 & 0.000 & 0.016 \\
5000 & 75 & 5/5 & 100.0 & 0.00000 & 12.79 & 90.3 & 76.0 & 135.5 & 0.000 & 0.038 \\
5000 & 100 & 5/5 & 100.0 & 0.00000 & 5.31 & 90.2 & 93.4 & 109.4 & 0.000 & 0.026 \\
6000 & 25 & 5/5 & 100.0 & 0.00000 & 19.36 & 108.5 & 97.4 & 204.4 & 0.000 & 0.050 \\
6000 & 50 & 5/5 & 100.0 & 0.00000 & 15.36 & 108.5 & 93.2 & 218.4 & 0.000 & 0.036 \\
6000 & 75 & 5/5 & 100.0 & -0.00156 & 11.37 & 108.6 & 74.4 & 226.3 & 0.002 & 0.046 \\
6000 & 100 & 5/5 & 100.0 & 0.00000 & 53.54 & 108.8 & 67.0 & 240.1 & 0.000 & 0.014 \\

\bottomrule
\end{tabular}
\end{adjustbox}
\end{table}

\subsection{Large-QKP}

For Large-QKP, we use the 120~s limit of \citet{Hochbaum2025} and compare with the best value obtained by any of the eight methods in their study, which in almost all cases is the value found by IHEA. Table~\ref{tab:large} shows that SSGA-P is never worse: it matches the reference value on 124 instances and improves it on 20. The improvements, listed in Table~\ref{tab:largenew}, concern graphs with at least 1000 items and mostly larger capacities. On 12 of them, all ten runs find the improved value. For $n=500$, where most reference values are proven optimal by exact solvers, every run of SSGA-P reaches them in less than 1.2~s.

The success rate decreases for the largest graph. With $n=10{,}000$ and capacities of 50 and 75\% of the total weight, a solution contains several thousand items, a generation costs about 0.2~s, and only about 500 generations fit into 120~s. Table~\ref{tab:n10000} compares runs of 120~s with runs of 600~s on this graph. With 600~s all 60 runs reach or exceed the reference value, and the value for the 10\% capacity improves further to 12{,}169{,}912.

\begin{table}[t]
\centering
\caption{Large-QKP, grouped by number of items: instances, instances on which SSGA-P is better than, equal to or worse than the best of the eight methods in \citet{Hochbaum2025}, percentage of SSGA-P runs reaching at least that value (10 runs, 120~s), and average gap (\%) of IHEA and QKBP in that study.}
\label{tab:large}
\small
\begin{adjustbox}{max width=\linewidth}
\begin{tabular}{rrrrrrrr}
\toprule
$n$ & Inst. & Better & Equal & Worse & Hits (\%) & IHEA gap & QKBP gap\\
\midrule
500 & 48 & 0 & 48 & 0 & 100.0 & 0.004 & 0.272 \\
1,000 & 36 & 4 & 32 & 0 & 99.4 & 0.000 & 0.107 \\
2,000 & 30 & 7 & 23 & 0 & 99.0 & 0.000 & 0.083 \\
5,000 & 24 & 6 & 18 & 0 & 96.2 & 0.000 & 0.028 \\
10,000 & 6 & 3 & 3 & 0 & 71.7 & 0.000 & 0.003 \\

\bottomrule
\end{tabular}
\end{adjustbox}
\end{table}

\begin{table}[t]
\centering
\caption{Large-QKP instances on which SSGA-P improves the best value reported by \citet{Hochbaum2025} (120~s, 10 runs; Runs is the number of runs reaching at least the reported value). Capacity $b$ is the index of the budget (1: 2.5\%, \dots, 6: 75\% of total weight).}
\label{tab:largenew}
\small
\begin{adjustbox}{max width=\linewidth}
\begin{tabular}{rrrrrrrr}
\toprule
$n$ & Dens. & $b$ & Best reported & SSGA-P & Difference & Runs & QKBP gap (\%)\\
\midrule
1,000 & 5 & 3 & 123,246 & \textbf{123,251} & +5 & 10/10 & 0.16 \\
1,000 & 5 & 4 & 317,319 & \textbf{317,340} & +21 & 10/10 & 0.05 \\
1,000 & 10 & 2 & 119,109 & \textbf{119,111} & +2 & 10/10 & 0.09 \\
1,000 & 20 & 5 & 2,537,611 & \textbf{2,537,666} & +55 & 10/10 & 0.18 \\
2,000 & 5 & 4 & 1,230,052 & \textbf{1,230,054} & +2 & 10/10 & 0.06 \\
2,000 & 5 & 6 & 3,789,519 & \textbf{3,789,520} & +1 & 10/10 & 0.02 \\
2,000 & 10 & 5 & 5,066,831 & \textbf{5,066,855} & +24 & 10/10 & 0.06 \\
2,000 & 10 & 6 & 7,625,072 & \textbf{7,625,205} & +133 & 10/10 & 0.04 \\
2,000 & 15 & 4 & 3,672,378 & \textbf{3,672,403} & +25 & 10/10 & 0.11 \\
2,000 & 20 & 5 & 9,944,933 & \textbf{9,945,193} & +260 & 10/10 & 0.08 \\
2,000 & 25 & 6 & 18,869,258 & \textbf{18,869,289} & +31 & 10/10 & 0.01 \\
5,000 & 5 & 4 & 7,774,287 & \textbf{7,774,293} & +6 & 10/10 & 0.00 \\
5,000 & 5 & 6 & 23,740,241 & \textbf{23,740,268} & +27 & 10/10 & 0.03 \\
5,000 & 10 & 5 & 31,583,225 & \textbf{31,583,249} & +24 & 10/10 & 0.04 \\
5,000 & 15 & 3 & 9,121,755 & \textbf{9,121,761} & +6 & 10/10 & 0.08 \\
5,000 & 15 & 6 & 70,862,287 & \textbf{70,862,324} & +37 & 10/10 & 0.03 \\
5,000 & 20 & 2 & 5,824,467 & \textbf{5,824,494} & +27 & 10/10 & 0.09 \\
10,000 & 5 & 1 & 2,838,130 & \textbf{2,838,163} & +33 & 10/10 & 0.01 \\
10,000 & 5 & 3 & 12,169,869 & \textbf{12,169,892} & +23 & 10/10 & 0.00 \\
10,000 & 5 & 4 & 30,848,899 & \textbf{30,848,924} & +25 & 7/10 & 0.01 \\

\bottomrule
\end{tabular}
\end{adjustbox}
\end{table}

\begin{table}[t]
\centering
\caption{Graph with $n=10{,}000$ items: best value and number of runs reaching the reference value after 120~s and after 600~s (10 runs). The 120~s values were recorded in the same runs when 120~s had elapsed, after applying the final descent to a copy of the best solution.}
\label{tab:n10000}
\small
\begin{adjustbox}{max width=\linewidth}
\begin{tabular}{lrrrrr}
\toprule
Capacity & Reference & Best (120 s) & Runs & Best (600 s) & Runs\\
\midrule
b1 (2.5\%) & 2,838,130 & \textbf{2,838,163} & 10/10 & \textbf{2,838,163} & 10/10 \\
b2 (5\%) & 5,922,475 & 5,922,475 & 9/10 & 5,922,475 & 10/10 \\
b3 (10\%) & 12,169,869 & \textbf{12,169,892} & 10/10 & \textbf{12,169,912} & 10/10 \\
b4 (25\%) & 30,848,899 & \textbf{30,848,924} & 7/10 & \textbf{30,848,924} & 10/10 \\
b5 (50\%) & 62,495,580 & 62,495,580 & 3/10 & 62,495,580 & 10/10 \\
b6 (75\%) & 94,402,497 & 94,402,497 & 3/10 & 94,402,497 & 10/10 \\

\bottomrule
\end{tabular}
\end{adjustbox}
\end{table}

\subsection{Comparison with QKBP in time}
\label{sec:qkbp}

QKBP is designed for speed and the 120~s limit is not a natural setting for it. To compare the two methods in time, we recorded in each run of SSGA-P, with a limit of 30~s, the first time at which the best solution was at least as good as the QKBP solution. The QKBP deviations in \citet{Hochbaum2025} are rounded to two decimals, so we used the largest QKBP value compatible with the rounding as the target. Table~\ref{tab:ttq} reports medians per graph size. All runs reached the target on 137 of the 144 instances; the seven exceptions are instances on which QKBP itself attains the reference value. SSGA-P needs a few tenths of a second for $n\le 1000$, about one second for $n=2000$ and 8 to 30~s for the largest graphs, which is typically 2 to 10 times the QKBP time including its parametric cut. QKBP is therefore clearly faster for a solution of its quality. After reaching the QKBP value, SSGA-P keeps improving and ends at or above the best value of all eight methods. The two methods therefore serve different purposes: QKBP gives a good solution almost immediately, and SSGA-P is preferable when the quality of the solution matters more than a few seconds or minutes of computing time.

\begin{table}[t]
\centering
\caption{Time to reach the QKBP solution value on Large-QKP (median over instances of the average over 10 runs, in seconds), compared with the running time of QKBP per capacity, the same time plus the parametric cut, and the running time of IHEA in \citet{Hochbaum2025}.}
\label{tab:ttq}
\small
\begin{adjustbox}{max width=\linewidth}
\begin{tabular}{rrrrrrr}
\toprule
$n$ & Inst. & All runs reach & SSGA-P & QKBP & QKBP + cut & IHEA\\
\midrule
500 & 48 & 48 & 0.06 & 0.01 & 0.04 & 6.0 \\
1,000 & 36 & 35 & 0.22 & 0.02 & 0.11 & 11.0 \\
2,000 & 30 & 30 & 1.05 & 0.10 & 0.31 & 42.1 \\
5,000 & 24 & 22 & 7.72 & 0.66 & 1.43 & 143.0 \\
10,000 & 6 & 2 & 29.73 & 2.78 & 4.31 & 205.1 \\

\bottomrule
\end{tabular}
\end{adjustbox}
\end{table}

\subsection{Ablation study}
\label{sec:ablation}

To measure the contribution of each component, we ran SSGA-P on Group~II with one component switched off at a time, with 10 runs per instance (Table~\ref{tab:ablation}). Removing the multi-ratio seeding has the largest effect: seven instances are no longer solved in any run, including 1000\_75\_5, 2000\_50\_5 and 2000\_100\_2, and the success rate falls from 99.8\% to 88.4\%. Removing the pair exchange leaves all BKVs reachable but lowers the success rate to 96.9\%, with ten instances losing runs, for example 1000\_50\_7 (4 of 10). Both effects are significant in a Wilcoxon test on the per-instance averages ($p=0.002$ and $p=0.005$). The effects of evict-and-insert and of the perturbation are not significant on Group~II. On five difficult Group~III instances (5000\_25\_5, 5000\_50\_4, 6000\_75\_1, 6000\_100\_2 and 6000\_100\_5), neither component changes the results either. For the perturbation, this follows from the setting: with $g_{\max}=n/2$ generations and fewer than 2000 generations per run, a kick is never triggered for $n\ge 5000$. On Large-QKP with $n\le 2000$, where 10 to 25 kicks occur per run, we repeated the test on the 24 instances of the graphs with $(n,\Delta)=(1000,5\%)$, $(1000,20\%)$, $(2000,10\%)$ and $(2000,20\%)$. Without evict-and-insert, and without the perturbation, all 240 runs still reach at least the reference value, and the six improved values of Table~\ref{tab:largenew} on these graphs are found in every run. The only difference in solution quality is on 2000\_10, budget 5, where the average value is 5{,}066{,}845.4 without evict-and-insert and 5{,}066{,}850.2 without the perturbation, against 5{,}066{,}852.6 for the complete algorithm. The average time to best on the 24 instances is 4.3~s for the complete algorithm, 5.6~s without evict-and-insert and 4.6~s without the perturbation. On the instances tested, these two components therefore act as safeguards rather than as sources of improvement, and we keep them because they cost little and do not harm the results.

\begin{table}[t]
\centering
\caption{Ablation on Group II: runs per instance, instances whose BKV is reached, instances with all runs successful, hits (\%), average gap (\%), time to best (s), number of instances where the complete algorithm has a better/worse average, and $p$-value of the Wilcoxon signed-rank test on per-instance averages.}
\label{tab:ablation}
\scriptsize
\setlength{\tabcolsep}{3.5pt}
\begin{adjustbox}{max width=\linewidth}
\begin{tabular}{lrrrrrrrr}
\toprule
Variant & Runs & BKV & All runs & Hits & Gap & $t_{\mathrm{best}}$ & Better/worse & $p$\\
\midrule
SSGA-P (all components) & 30 & 80 & 78 & 99.79 & 0.000000 & 1.41 & -- & -- \\
SSGA-P without multi-ratio seeding & 10 & 73 & 68 & 88.38 & 0.002572 & 2.50 & 12/1 & 0.002 \\
SSGA-P without pair exchange & 10 & 80 & 70 & 96.88 & 0.000010 & 1.55 & 10/0 & 0.005 \\
SSGA-P without evict-and-insert & 10 & 80 & 77 & 99.50 & 0.000002 & 1.54 & 2/1 & 0.285 \\
SSGA-P without perturbation & 10 & 80 & 79 & 99.88 & 0.000000 & 1.57 & 0/2 & 0.180 \\

\bottomrule
\end{tabular}
\end{adjustbox}
\end{table}

\begin{figure}[t]
\centering
\begin{adjustbox}{max width=\linewidth}
\begin{tikzpicture}
\begin{axis}[width=6cm,height=5cm,xlabel={CPU time (s)},ylabel={average gap (\%)},title={2000\_50\_5 (Group II)},xmin=0,xmax=14,ymin=-0.0004,ymax=0.009,grid=both,grid style={gray!20},unbounded coords=jump,tick label style={font=\scriptsize},label style={font=\small},title style={font=\small},y tick label style={/pgf/number format/fixed,/pgf/number format/precision=4},scaled y ticks=false,]
\addplot[blue!70,line width=2.4pt,const plot] table[x=t,y=full]{figures/conv_g2.dat};
\addplot[red,densely dashed,thick,const plot] table[x=t,y=nobeta]{figures/conv_g2.dat};
\addplot[green!50!black,densely dotted,line width=1.2pt,const plot] table[x=t,y=nopair]{figures/conv_g2.dat};
\draw[gray,dashdotted] ({axis cs:12,0}|-{rel axis cs:0,0}) -- ({axis cs:12,0}|-{rel axis cs:0,1});
\addplot[black,dashed,domain=0:14] {0.00496};
\end{axis}
\end{tikzpicture}\hspace{2mm}%
\begin{tikzpicture}
\begin{axis}[width=6cm,height=5cm,xlabel={CPU time (s)},title={5000\_50\_4 (Group III)},xmin=0,xmax=100,ymin=-0.00005,ymax=0.0012,grid=both,grid style={gray!20},unbounded coords=jump,tick label style={font=\scriptsize},label style={font=\small},title style={font=\small},y tick label style={/pgf/number format/fixed,/pgf/number format/precision=4},scaled y ticks=false,]
\addplot[blue!70,line width=2.4pt,const plot] table[x=t,y=full]{figures/conv_g3.dat};
\addplot[red,densely dashed,thick,const plot] table[x=t,y=nobeta]{figures/conv_g3.dat};
\addplot[green!50!black,densely dotted,line width=1.2pt,const plot] table[x=t,y=nopair]{figures/conv_g3.dat};
\draw[gray,dashdotted] ({axis cs:90,0}|-{rel axis cs:0,0}) -- ({axis cs:90,0}|-{rel axis cs:0,1});
\end{axis}
\end{tikzpicture}\hspace{2mm}%
\begin{tikzpicture}
\begin{axis}[width=6cm,height=5cm,xlabel={CPU time (s)},title={Large-QKP 2000, 20\%, $b=5$},legend style={font=\scriptsize,at={(0.97,0.97)},anchor=north east},xmin=0,xmax=130,ymin=-0.00035,ymax=0.008,grid=both,grid style={gray!20},unbounded coords=jump,tick label style={font=\scriptsize},label style={font=\small},title style={font=\small},y tick label style={/pgf/number format/fixed,/pgf/number format/precision=4},scaled y ticks=false,]
\addplot[blue!70,line width=2.4pt,const plot] table[x=t,y=full]{figures/conv_lq.dat};
\addplot[red,densely dashed,thick,const plot] table[x=t,y=nobeta]{figures/conv_lq.dat};
\addplot[green!50!black,densely dotted,line width=1.2pt,const plot] table[x=t,y=nopair]{figures/conv_lq.dat};
\draw[gray,dashdotted] ({axis cs:120,0}|-{rel axis cs:0,0}) -- ({axis cs:120,0}|-{rel axis cs:0,1});
\addplot[black,dashed,domain=0:130] {0.00261};
\legend{SSGA-P,no $\beta$ grid,no pair exchange}
\end{axis}
\end{tikzpicture}
\end{adjustbox}
\caption{Average gap (\%) over 10 runs between the best value found so far and the best known value, as a function of CPU time, for SSGA-P and the variants without multi-ratio seeding and without pair exchange. For the Large-QKP instance the reference is the improved value 9{,}945{,}193. Dashed black lines show the best value obtained in the study of \citet{Hochbaum2025} (gap 0.0050\% and 0.0026\%). The QKBP values lie outside the plotted range (gaps of about 0.1\% and 0.08\%). Curves start once every run has completed its initial population, so the empty part on the left is the time needed to build it. Where curves coincide at zero gap (for example, the complete algorithm and the variant without pair exchange on 2000\_50\_5), the dotted line is drawn on top of the thick line; the horizontal axis is placed slightly below zero so that these curves remain visible. The vertical dash-dotted line marks the time limit, after which the final values are shown; the vertical axes are cut off so that the final phase is visible, and the early values of the variant without seeding lie above them.}
\label{fig:conv}
\end{figure}
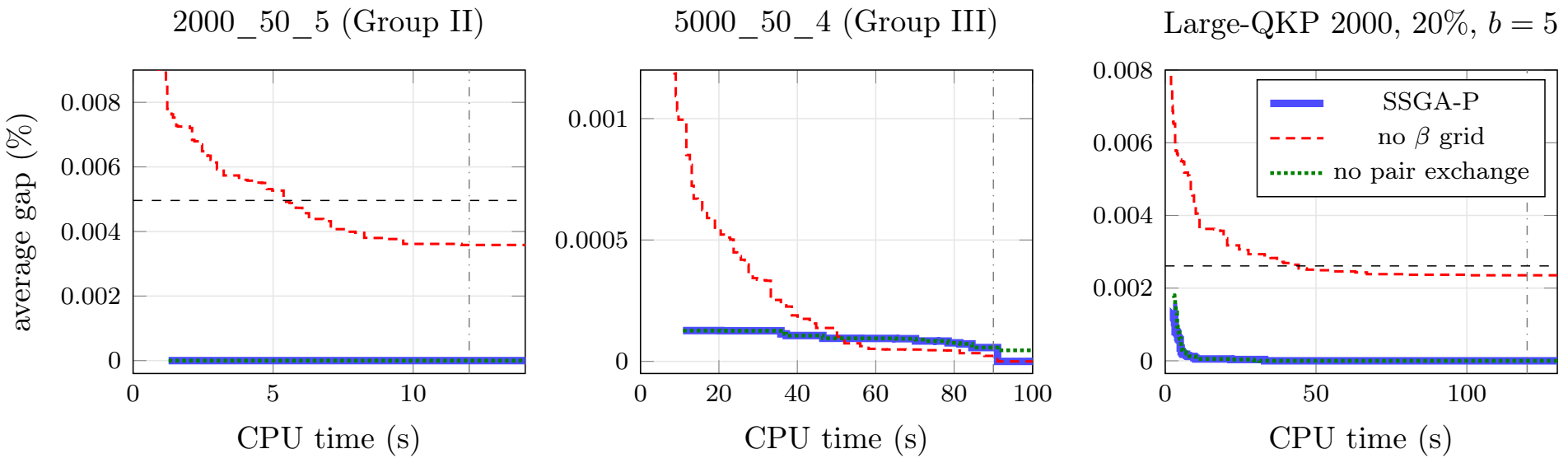

Figure~\ref{fig:conv} shows how the search progresses on three instances. On 2000\_50\_5 the multi-ratio seeding places the best known value 16{,}806{,}059 in the initial population in all ten runs, so the best solution is found after about 1.2~s, before the first generation; without it, the runs improve slowly and all of them stop at least 0.003\% below that value. On 5000\_50\_4, the instance of Section~\ref{sec:analysis}, the variant without pair exchange approaches the best known value at the same speed as the complete algorithm but 3 of its 10 runs remain at a budget-tight local optimum until the time limit. On this instance, the seeding brings no benefit: the variant without it improves more slowly at first but is slightly ahead after about 50~s, and the complete algorithm reaches the best known value in all runs only through the final descent at the time limit. Since no kick occurs for $n\ge 5000$ within the time limit, the pair exchange is applied on this instance only once, in the final descent, and the figure shows that this single application closes the remaining gap. On the Large-QKP instance the complete algorithm passes the best value of the 2025 study within a few seconds and reaches 9{,}945{,}193 in every run after 10.5~s on average, while the variant without seeding needs 50~s on average and has a larger spread. The figure also shows the price of the seeding: building and improving the 51 greedy solutions delays the first complete population, from 0.3~s to 1.3~s for $n=2000$ and from 2.4~s to 10.6~s for $n=5000$. On the instances where the seeding helps, this cost is small compared with the time it saves; where it does not help, as on 5000\_50\_4, it is a few seconds out of 90.

\subsection{Discussion}

The results suggest that the classical QKP benchmarks are less settled than their age would indicate. IHEA solves the instances with up to 2000 items almost perfectly, but on the largest standard instances its success rate drops to 87.7\%, and on Large-QKP the best values of eight methods could be improved on 20 of 96 instances with at least 1000 items. In the instances we analysed, the remaining difficulty is not the size of the search space as such but a mismatch between ratio-based rankings and the item mix of the best solutions, and the presence of local optima that only coupled exchanges can leave. Both are cheap to address: the seeding costs 51 greedy constructions and the pair exchange is used only a few times per run.

The method has limitations. It is slower than QKBP for comparable solution quality and, unlike QKBP, it solves each capacity separately. The time limit for the largest instances had to be increased relative to the smaller ones, and for $n=10{,}000$ with large capacities 120~s is not enough for all runs to reach the best values. Finally, the perturbation does not contribute on the largest instances with the present trigger, and the evict-and-insert move is largely redundant once the other two components are present.

\section{Conclusions}
\label{sec:concl}

We studied why ratio-based search for the quadratic knapsack problem fails on some benchmark instances and identified two causes shared by most QKP heuristics: the bias of the profit-to-weight ranking, which can exclude the item mix of the best solutions, and local optima on almost full knapsacks that can only be left by two coupled exchanges. Our proposed approach, namely SSGA-P, addresses them with a multi-ratio seeding of the initial population and a pair exchange move inside a simple steady-state genetic algorithm. Our approach reaches all best known values of the classical collections with up to 6000 items, is significantly more reliable than IHEA \citep{ChenHao2017} on the largest of them (99.75\% against 87.7\% of the runs), improves one best known value, and improves the best values of the eight-method comparison of \citet{Hochbaum2025} on 20 of the 144 Large-QKP instances without being worse on any. The ablation study shows that these gains come from the two new components: without the seeding, the success rate on Group~II falls from 99.79\% to 88.4\%, and without the pair exchange to 96.9\%. QKBP \citep{Hochbaum2025} remains faster for solutions of slightly lower quality. SSGA-P not only provides solutions of superior  quality, but it is  much simpler approach in comparison to IHEA and QKBP which facilitates ease of understanding and implementation.

Several directions remain open. The exponent $\beta$ could be adapted during the search instead of being sampled on a grid, and the pair exchange could be extended to longer ejection chains. Using breakpoint solutions of QKBP as seeds may combine speed and accuracy.  More broadly, the two components are not tied to the genetic algorithm and could be inserted into other metaheuristics, such as tabu search or iterated local search. They also extend naturally to related problems with pairwise profits and capacity constraints, including the quadratic multiple knapsack problem, the multidimensional and multiple-choice variants of the QKP, and budgeted selection problems arising in applications such as facility location and sensor placement, where item rankings by value per unit of cost are equally common.

\section*{Data and code availability}
The benchmark instances used in this study are publicly available from the sources cited in Section~\ref{sec:exp}. The source code of SSGA-P, the detailed results of all runs and the improved solutions reported in this paper will be made available by the corresponding author upon reasonable request.

\section*{CRediT authorship contribution statement}
\textbf{Gaurav Srivastava:} Conceptualization, Methodology, Software, Validation, Formal analysis, Investigation, Visualization, Writing -- review \& editing, Supervision.

\textbf{Rahul Verma:} Software, Investigation, Visualization, Writing -- original draft.

\textbf{Alok Singh:} Conceptualization, Validation, Formal analysis, Writing -- review \& editing, Supervision.

\section*{Declaration of competing interest}
The authors declare that they have no known competing financial interests or personal relationships that could have appeared to influence the work reported in this paper.

\section*{Acknowledgements}
The first and second authors thank the Indian Institute of Information Technology Allahabad, Prayagraj, for providing the research and computing facilities that supported this work.

\bibliographystyle{elsarticle-harv}
\bibliography{refs}

\appendix
\section{Per-instance results}
\label{app:inst}

Tables~\ref{tab:appg2} and~\ref{tab:appg3} list the results per instance. Bold values improve the BKV. SR is the IHEA success rate (\%) and $t_b$ the IHEA time to best (s) from \citet{ChenHao2017}; the QKBP gap (\%) is taken from \citet{Hochbaum2025}.

\scriptsize
\setlength{\tabcolsep}{2pt}
\begin{longtable}{lrrrrr rrr r}
\caption{Group II, per instance (SSGA-P: 30 runs).}\label{tab:appg2}\\
\toprule
Instance & BKV & Best & Average & Hits & $t_{\mathrm{best}}$ & IHEA SR & IHEA $t_b$ & GRASP SR & QKBP gap\\
\midrule
\endfirsthead
\toprule
Instance & BKV & Best & Average & Hits & $t_{\mathrm{best}}$ & IHEA SR & IHEA $t_b$ & GRASP SR & QKBP gap\\
\midrule
\endhead
1000\_25\_1 & 6,172,407 & 6,172,407 & 6,172,407.0 & 30 & 0.62 & 100 & 0.09 & 100 & 0.16 \\
1000\_25\_2 & 229,941 & 229,941 & 229,941.0 & 30 & 0.22 & 100 & 0.09 & 66 & 0.00 \\
1000\_25\_3 & 172,418 & 172,418 & 172,418.0 & 30 & 0.11 & 100 & 0.05 & 100 & 0.04 \\
1000\_25\_4 & 367,426 & 367,426 & 367,426.0 & 30 & 0.14 & 100 & 0.07 & 100 & 0.00 \\
1000\_25\_5 & 4,885,611 & 4,885,611 & 4,885,611.0 & 30 & 1.73 & 100 & 0.14 & 100 & 0.15 \\
1000\_25\_6 & 15,689 & 15,689 & 15,689.0 & 30 & 0.04 & 100 & 0.02 & 100 & 1.27 \\
1000\_25\_7 & 4,945,810 & 4,945,810 & 4,945,810.0 & 30 & 0.72 & 100 & 0.31 & 100 & 0.13 \\
1000\_25\_8 & 1,710,198 & 1,710,198 & 1,710,198.0 & 30 & 0.70 & 100 & 0.42 & 100 & 0.10 \\
1000\_25\_9 & 496,315 & 496,315 & 496,315.0 & 30 & 0.16 & 100 & 0.06 & 100 & 0.00 \\
1000\_25\_10 & 1,173,792 & 1,173,792 & 1,173,792.0 & 30 & 0.30 & 100 & 0.15 & 100 & 0.24 \\
1000\_50\_1 & 5,663,590 & 5,663,590 & 5,663,590.0 & 30 & 0.37 & 100 & 0.11 & 100 & 0.02 \\
1000\_50\_2 & 180,831 & 180,831 & 180,831.0 & 30 & 0.08 & 100 & 0.04 & 100 & 0.00 \\
1000\_50\_3 & 11,384,283 & 11,384,283 & 11,384,283.0 & 30 & 0.57 & 100 & 0.09 & 100 & 0.01 \\
1000\_50\_4 & 322,226 & 322,226 & 322,226.0 & 30 & 0.11 & 100 & 0.06 & 100 & 0.11 \\
1000\_50\_5 & 9,984,247 & 9,984,247 & 9,984,246.1 & 29 & 1.87 & 98 & 0.54 & 86 & 0.08 \\
1000\_50\_6 & 4,106,261 & 4,106,261 & 4,106,261.0 & 30 & 0.31 & 100 & 0.11 & 100 & 0.12 \\
1000\_50\_7 & 10,498,370 & 10,498,370 & 10,498,370.0 & 30 & 4.57 & 100 & 0.27 & 84 & 0.03 \\
1000\_50\_8 & 4,981,146 & 4,981,146 & 4,981,145.1 & 26 & 1.36 & 99 & 1.65 & 20 & 0.14 \\
1000\_50\_9 & 1,727,861 & 1,727,861 & 1,727,861.0 & 30 & 0.22 & 100 & 0.85 & 100 & 0.00 \\
1000\_50\_10 & 2,340,724 & 2,340,724 & 2,340,724.0 & 30 & 0.90 & 100 & 0.16 & 94 & 0.21 \\
1000\_75\_1 & 11,570,056 & 11,570,056 & 11,570,056.0 & 30 & 3.63 & 100 & 0.51 & 65 & 0.08 \\
1000\_75\_2 & 1,901,389 & 1,901,389 & 1,901,389.0 & 30 & 0.18 & 100 & 0.13 & 100 & 0.13 \\
1000\_75\_3 & 2,096,485 & 2,096,485 & 2,096,485.0 & 30 & 0.18 & 100 & 0.11 & 100 & 0.54 \\
1000\_75\_4 & 7,305,321 & 7,305,321 & 7,305,321.0 & 30 & 0.38 & 100 & 0.12 & 100 & 0.09 \\
1000\_75\_5 & 13,970,842 & 13,970,842 & 13,970,842.0 & 30 & 0.40 & 100 & 0.13 & 93 & 0.14 \\
1000\_75\_6 & 12,288,738 & 12,288,738 & 12,288,738.0 & 30 & 0.39 & 100 & 0.16 & 100 & 0.06 \\
1000\_75\_7 & 1,095,837 & 1,095,837 & 1,095,837.0 & 30 & 0.14 & 100 & 0.10 & 100 & 0.69 \\
1000\_75\_8 & 5,575,813 & 5,575,813 & 5,575,813.0 & 30 & 0.27 & 100 & 0.14 & 100 & 0.12 \\
1000\_75\_9 & 695,774 & 695,774 & 695,774.0 & 30 & 0.11 & 100 & 0.13 & 100 & 0.42 \\
1000\_75\_10 & 2,507,677 & 2,507,677 & 2,507,677.0 & 30 & 0.20 & 100 & 0.07 & 100 & 0.05 \\
1000\_100\_1 & 6,243,494 & 6,243,494 & 6,243,494.0 & 30 & 0.45 & 100 & 0.12 & 100 & 0.13 \\
1000\_100\_2 & 4,854,086 & 4,854,086 & 4,854,086.0 & 30 & 1.26 & 100 & 0.19 & 61 & 0.22 \\
1000\_100\_3 & 3,172,022 & 3,172,022 & 3,172,022.0 & 30 & 0.28 & 100 & 0.10 & 100 & 0.28 \\
1000\_100\_4 & 754,727 & 754,727 & 754,727.0 & 30 & 0.16 & 100 & 0.07 & 100 & 0.52 \\
1000\_100\_5 & 18,646,620 & 18,646,620 & 18,646,620.0 & 30 & 2.98 & 100 & 0.29 & 99 & 0.08 \\
1000\_100\_6 & 16,020,232 & 16,020,232 & 16,020,232.0 & 30 & 0.43 & 100 & 0.12 & 96 & 0.09 \\
1000\_100\_7 & 12,936,205 & 12,936,205 & 12,936,205.0 & 30 & 0.47 & 100 & 0.13 & 100 & 0.03 \\
1000\_100\_8 & 6,927,738 & 6,927,738 & 6,927,738.0 & 30 & 1.58 & 100 & 0.11 & 100 & 0.11 \\
1000\_100\_9 & 3,874,959 & 3,874,959 & 3,874,959.0 & 30 & 0.23 & 100 & 0.07 & 100 & 0.02 \\
1000\_100\_10 & 1,334,494 & 1,334,494 & 1,334,494.0 & 30 & 0.21 & 100 & 0.09 & 100 & 0.36 \\
2000\_25\_1 & 5,268,188 & 5,268,188 & 5,268,188.0 & 30 & 1.99 & 100 & 0.48 & 100 & 0.00 \\
2000\_25\_2 & 13,294,030 & 13,294,030 & 13,294,030.0 & 30 & 1.52 & 100 & 2.19 & 100 & 0.11 \\
2000\_25\_3 & 5,500,433 & 5,500,433 & 5,500,433.0 & 30 & 2.31 & 100 & 0.78 & 56 & 0.12 \\
2000\_25\_4 & 14,625,118 & 14,625,118 & 14,625,118.0 & 30 & 4.54 & 100 & 0.77 & 100 & 0.01 \\
2000\_25\_5 & 5,975,751 & 5,975,751 & 5,975,751.0 & 30 & 1.74 & 100 & 0.50 & 100 & 0.15 \\
2000\_25\_6 & 4,491,691 & 4,491,691 & 4,491,691.0 & 30 & 0.93 & 100 & 0.77 & 100 & 0.00 \\
2000\_25\_7 & 6,388,756 & 6,388,756 & 6,388,756.0 & 30 & 1.42 & 100 & 0.67 & 100 & 0.01 \\
2000\_25\_8 & 11,769,873 & 11,769,873 & 11,769,873.0 & 30 & 1.44 & 100 & 0.72 & 100 & 0.00 \\
2000\_25\_9 & 10,960,328 & 10,960,328 & 10,960,328.0 & 30 & 3.83 & 100 & 0.56 & 100 & 0.00 \\
2000\_25\_10 & 139,236 & 139,236 & 139,236.0 & 30 & 0.22 & 100 & 0.09 & 100 & 0.00 \\
2000\_50\_1 & 7,070,736 & 7,070,736 & 7,070,736.0 & 30 & 0.85 & 100 & 0.44 & 39 & 0.19 \\
2000\_50\_2 & 12,587,545 & 12,587,545 & 12,587,545.0 & 30 & 1.69 & 100 & 1.35 & 100 & 0.05 \\
2000\_50\_3 & 27,268,336 & 27,268,336 & 27,268,336.0 & 30 & 1.57 & 100 & 0.36 & 100 & 0.00 \\
2000\_50\_4 & 17,754,434 & 17,754,434 & 17,754,434.0 & 30 & 7.44 & 100 & 0.45 & 100 & 0.08 \\
2000\_50\_5 & 16,806,059 & 16,806,059 & 16,806,059.0 & 30 & 1.26 & 100 & 0.54 & 90 & 0.10 \\
2000\_50\_6 & 23,076,155 & 23,076,155 & 23,076,155.0 & 30 & 3.72 & 100 & 0.63 & 50 & 0.05 \\
2000\_50\_7 & 28,759,759 & 28,759,759 & 28,759,759.0 & 30 & 1.59 & 100 & 0.53 & 6 & 0.08 \\
2000\_50\_8 & 1,580,242 & 1,580,242 & 1,580,242.0 & 30 & 0.47 & 100 & 0.17 & 100 & 0.12 \\
2000\_50\_9 & 26,523,791 & 26,523,791 & 26,523,791.0 & 30 & 1.55 & 100 & 0.44 & 100 & 0.04 \\
2000\_50\_10 & 24,747,047 & 24,747,047 & 24,747,047.0 & 30 & 1.48 & 100 & 0.38 & 100 & 0.00 \\
2000\_75\_1 & 25,121,998 & 25,121,998 & 25,121,998.0 & 30 & 1.23 & 100 & 0.53 & 100 & 0.06 \\
2000\_75\_2 & 12,664,670 & 12,664,670 & 12,664,670.0 & 30 & 0.92 & 100 & 0.40 & 89 & 0.08 \\
2000\_75\_3 & 43,943,994 & 43,943,994 & 43,943,994.0 & 30 & 1.61 & 100 & 0.76 & 100 & 0.02 \\
2000\_75\_4 & 37,496,613 & 37,496,613 & 37,496,613.0 & 30 & 2.41 & 100 & 0.43 & 100 & 0.03 \\
2000\_75\_5 & 24,835,349 & 24,835,349 & 24,835,349.0 & 30 & 1.22 & 100 & 0.53 & 73 & 0.10 \\
2000\_75\_6 & 45,137,758 & 45,137,758 & 45,137,758.0 & 30 & 1.64 & 100 & 0.34 & 100 & 0.00 \\
2000\_75\_7 & 25,502,608 & 25,502,608 & 25,502,608.0 & 30 & 1.24 & 100 & 0.40 & 100 & 0.03 \\
2000\_75\_8 & 10,067,892 & 10,067,892 & 10,067,892.0 & 30 & 2.59 & 100 & 0.33 & 100 & 0.04 \\
2000\_75\_9 & 14,177,079 & 14,177,079 & 14,177,079.0 & 30 & 0.94 & 100 & 0.48 & 97 & 0.16 \\
2000\_75\_10 & 7,815,755 & 7,815,755 & 7,815,755.0 & 30 & 1.38 & 100 & 1.73 & 78 & 0.14 \\
2000\_100\_1 & 37,929,909 & 37,929,909 & 37,929,909.0 & 30 & 1.32 & 100 & 0.42 & 100 & 0.02 \\
2000\_100\_2 & 33,665,281 & 33,665,281 & 33,665,281.0 & 30 & 1.23 & 100 & 0.55 & 95 & 0.08 \\
2000\_100\_3 & 29,952,019 & 29,952,019 & 29,952,019.0 & 30 & 6.27 & 100 & 0.44 & 34 & 0.06 \\
2000\_100\_4 & 26,949,268 & 26,949,268 & 26,949,268.0 & 30 & 2.21 & 100 & 0.54 & 100 & 0.05 \\
2000\_100\_5 & 22,041,715 & 22,041,715 & 22,041,715.0 & 30 & 9.19 & 95 & 6.29 & 70 & 0.11 \\
2000\_100\_6 & 18,868,887 & 18,868,887 & 18,868,887.0 & 30 & 0.96 & 100 & 0.39 & 100 & 0.06 \\
2000\_100\_7 & 15,850,597 & 15,850,597 & 15,850,597.0 & 30 & 2.04 & 100 & 0.40 & 100 & 0.07 \\
2000\_100\_8 & 13,628,967 & 13,628,967 & 13,628,967.0 & 30 & 0.84 & 100 & 0.36 & 100 & 0.10 \\
2000\_100\_9 & 8,394,562 & 8,394,562 & 8,394,562.0 & 30 & 1.44 & 100 & 0.36 & 97 & 0.15 \\
2000\_100\_10 & 4,923,559 & 4,923,559 & 4,923,559.0 & 30 & 1.56 & 100 & 0.91 & 92 & 0.21 \\

\bottomrule
\end{longtable}

\begin{longtable}{lrrrrr rr r}
\caption{Group III, per instance (SSGA-P: 30 runs).}\label{tab:appg3}\\
\toprule
Instance & BKV & Best & Average & Hits & $t_{\mathrm{best}}$ & IHEA SR & IHEA $t_b$ & QKBP gap\\
\midrule
\endfirsthead
\toprule
Instance & BKV & Best & Average & Hits & $t_{\mathrm{best}}$ & IHEA SR & IHEA $t_b$ & QKBP gap\\
\midrule
\endhead
5000\_25\_1 & 23,667,450 & 23,667,450 & 23,667,450.0 & 30 & 6.16 & 100 & 1.97 & 0.01 \\
5000\_25\_2 & 37,914,560 & 37,914,560 & 37,914,560.0 & 30 & 22.51 & 100 & 2.84 & 0.03 \\
5000\_25\_3 & 68,295,820 & 68,295,820 & 68,295,820.0 & 30 & 13.87 & 100 & 2.67 & 0.00 \\
5000\_25\_4 & 33,866,053 & 33,866,053 & 33,866,053.0 & 30 & 11.37 & 100 & 3.37 & 0.02 \\
5000\_25\_5 & 9,533,115 & 9,533,115 & 9,533,105.1 & 27 & 21.50 & 100 & 3.23 & 0.08 \\
5000\_50\_1 & 45,194,685 & 45,194,685 & 45,194,685.0 & 30 & 41.93 & 100 & 2.50 & 0.03 \\
5000\_50\_2 & 88,355,678 & 88,355,678 & 88,355,678.0 & 30 & 9.46 & 100 & 2.49 & 0.01 \\
5000\_50\_3 & 152,447,303 & 152,447,303 & 152,447,303.0 & 30 & 78.04 & 100 & 2.96 & 0.02 \\
5000\_50\_4 & 171,000,228 & 171,000,228 & 171,000,228.0 & 30 & 59.30 & 100 & 3.39 & 0.02 \\
5000\_50\_5 & 1,187,339 & 1,187,339 & 1,187,339.0 & 30 & 0.64 & 100 & 0.33 & 0.00 \\
5000\_75\_1 & 28,170,819 & 28,170,819 & 28,170,819.0 & 30 & 3.22 & 100 & 1.67 & 0.07 \\
5000\_75\_2 & 195,434,758 & 195,434,758 & 195,434,758.0 & 30 & 19.56 & 100 & 2.83 & 0.01 \\
5000\_75\_3 & 64,324,704 & 64,324,704 & 64,324,704.0 & 30 & 5.78 & 76 & 36.11 & 0.04 \\
5000\_75\_4 & 247,348,595 & 247,348,595 & 247,348,595.0 & 30 & 23.66 & 100 & 2.94 & 0.00 \\
5000\_75\_5 & 46,462,750 & 46,462,750 & 46,462,750.0 & 30 & 11.73 & 4 & 45.05 & 0.07 \\
5000\_100\_1 & 214,425,886 & 214,425,886 & 214,425,886.0 & 30 & 9.66 & 100 & 2.97 & 0.02 \\
5000\_100\_2 & 18,783,132 & 18,783,132 & 18,783,132.0 & 30 & 2.58 & 100 & 1.16 & 0.05 \\
5000\_100\_3 & 10,784,650 & 10,784,650 & 10,784,650.0 & 30 & 1.38 & 100 & 0.65 & 0.00 \\
5000\_100\_4 & 160,539,947 & 160,539,947 & 160,539,947.0 & 30 & 9.19 & 100 & 2.78 & 0.02 \\
5000\_100\_5 & 33,166,524 & 33,166,524 & 33,166,524.0 & 30 & 3.72 & 67 & 38.98 & 0.04 \\
6000\_25\_1 & 69,832,542 & 69,832,542 & 69,832,542.0 & 30 & 18.27 & 100 & 5.62 & 0.00 \\
6000\_25\_2 & 3,697,236 & 3,697,236 & 3,697,236.0 & 30 & 2.27 & 100 & 2.80 & 0.20 \\
6000\_25\_3 & 79,300,092 & 79,300,092 & 79,300,092.0 & 30 & 14.84 & 100 & 3.49 & 0.00 \\
6000\_25\_4 & 191,531,304 & 191,531,304 & 191,531,304.0 & 30 & 43.00 & 87 & 9.09 & 0.02 \\
6000\_25\_5 & 36,121,510 & 36,121,510 & 36,121,510.0 & 30 & 18.43 & 100 & 3.45 & 0.03 \\
6000\_50\_1 & 194,344,567 & 194,344,567 & 194,344,567.0 & 30 & 21.38 & 100 & 4.06 & 0.00 \\
6000\_50\_2 & 323,753,804 & 323,753,804 & 323,753,804.0 & 30 & 15.36 & 66 & 4.86 & 0.02 \\
6000\_50\_3 & 31,913,824 & 31,913,824 & 31,913,824.0 & 30 & 5.51 & 100 & 3.02 & 0.11 \\
6000\_50\_4 & 225,556,641 & 225,556,641 & 225,556,641.0 & 30 & 23.24 & 100 & 5.78 & 0.00 \\
6000\_50\_5 & 40,931,924 & 40,931,924 & 40,931,924.0 & 30 & 11.29 & 100 & 6.85 & 0.05 \\
6000\_75\_1 & 204,512,250 & \textbf{204,528,201} & 204,528,201.0 & 30 & 14.18 & 12 & 86.72 & 0.03 \\
6000\_75\_2 & 42,422,207 & 42,422,207 & 42,422,207.0 & 30 & 5.13 & 100 & 2.75 & 0.07 \\
6000\_75\_3 & 524,508,156 & 524,508,156 & 524,508,156.0 & 30 & 15.58 & 60 & 24.38 & 0.03 \\
6000\_75\_4 & 197,004,931 & 197,004,931 & 197,004,931.0 & 30 & 14.51 & 100 & 4.10 & 0.01 \\
6000\_75\_5 & 74,350,712 & 74,350,712 & 74,350,712.0 & 30 & 7.43 & 100 & 3.82 & 0.08 \\
6000\_100\_1 & 292,257,056 & 292,257,056 & 292,257,056.0 & 30 & 21.90 & 100 & 4.05 & 0.01 \\
6000\_100\_2 & 219,791,358 & 219,791,358 & 219,791,358.0 & 30 & 109.08 & 1 & 122.76 & 0.03 \\
6000\_100\_3 & 376,967,122 & 376,967,122 & 376,967,122.0 & 30 & 14.87 & 96 & 46.10 & 0.02 \\
6000\_100\_4 & 355,609,720 & 355,609,720 & 355,609,720.0 & 30 & 12.78 & 100 & 3.71 & 0.00 \\
6000\_100\_5 & 686,364,195 & 686,364,195 & 686,364,195.0 & 30 & 109.09 & 38 & 69.90 & 0.01 \\

\bottomrule
\end{longtable}
\normalsize

\end{document}